\documentclass{ieeeaccess}
\usepackage{cite}
\usepackage{amsmath,amssymb,amsfonts}
\usepackage{algorithmic}
\usepackage{graphicx}
\usepackage{textcomp}
\usepackage[ruled,vlined]{algorithm2e}

\begin{document}
\history{Date of publication xxxx 00, 0000, date of current version xxxx 00, 0000.}
\doi{10.1109/ACCESS.2017.DOI}

\title{Height Prediction and Refinement from Aerial Images with Semantic and Geometric Guidance}

\author{Mahdi Elhousni}
%\email{melhousni@wpi.edu}
%\affiliation{%
%  \institution{Worcester Polytechnic Institute}
%}

\author{Ziming Zhang}
%\email{zzhang@wpi.edu}
%\affiliation{%
%  \institution{Worcester Polytechnic Institute}
%}

\author{Xinming Huang}
%\email{xhuang@wpi.edu}
%\affiliation{%
%  \institution{Worcester Polytechnic Institute}
%}

\author{\uppercase{Mahdi Elhousni},
\uppercase{Ziming Zhang and Xinming Huang}.}
\address{Department of Electrical and Computer Engineering, Worcester Polytechnic Institute, Worcester, MA 01609, USA}

\corresp{Corresponding author: Xinming Huang (e-mail: xhuang@wpi.edu).}

\begin{abstract}
Deep learning provides a powerful new approach to many computer vision tasks. Height prediction from aerial images is one of those tasks which benefited greatly from the deployment of deep learning, thus replacing traditional multi-view geometry techniques. This manuscript proposes a two-stage approach to solve this task, where the first stage is a multi-task neural network whose main branch is used to predict the height map resulting from a single RGB aerial input image, while being augmented with semantic and geometric information from two additional branches. The second stage is a refinement step, where a denoising autoencoder is used to correct some errors in the first stage prediction results, producing a more accurate height map. Experiments on two publicly available datasets show that the proposed method is able to outperform state-of-the-art computer vision based and deep learning-based height prediction methods. 
Code is publicly available at : https://github.com/melhousni/DSMNet.
\end{abstract}

\begin{keywords}
UAV, Height, DSM, CNN, Autoencoders, Multi-Task.
\end{keywords}

\titlepgskip=-15pt

\maketitle

\section{Introduction}
\IEEEPARstart{A}{erial} imagery analysis was known as a very tedious task owing to the low quality of the acquired images and the lack of some appropriate automated process that could extract the relevant information from the data. Fortunately, recent advances in computer vision have made it possible to directly extract predefined patterns from the images, by applying some carefully designed algorithms. Moreover, deep learning brings in a new revolution to the field of aerial imagery analysis with more intelligence and better accuracy. As a result, multiple deep learning challenges related to aerial imagery processing, such as semantic segmentation \cite{chen2018semantic,marmanis2016semantic} and object detection \cite{ding2019learning,vsevo2016convolutional}, have been routinely featured each year by the geoscience and remote sensing (GRSS) community \cite{le20182018},\cite{le20192019},\cite{yokoya20202020}.  \par

\begin{figure}[t]
    \begin{center}
        \includegraphics[width=0.99\linewidth]{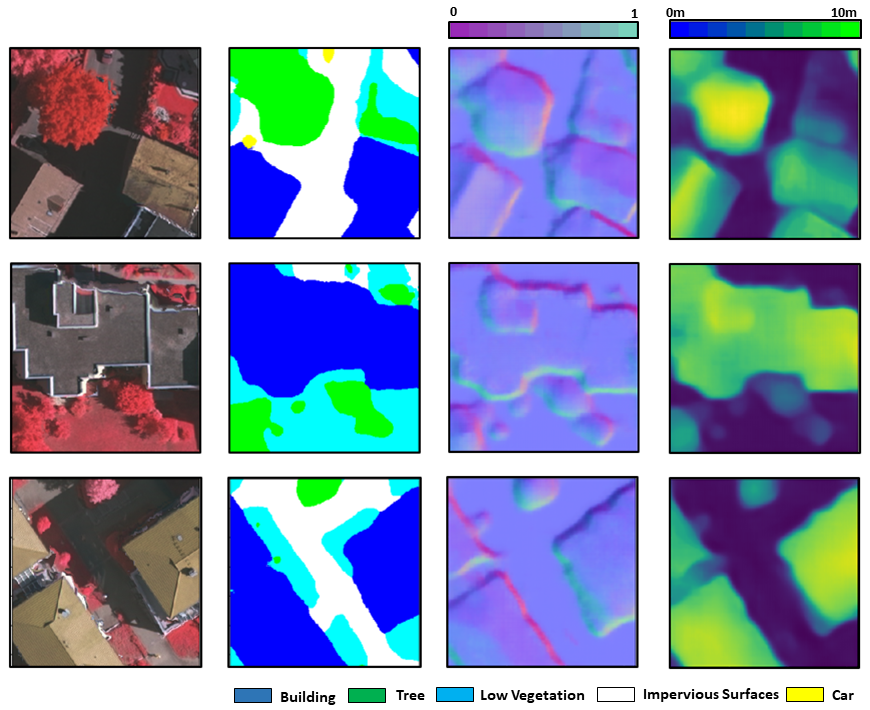}
        \caption{The outputs of our multi-task network. From left to right: The input RGB image, the output semantic labels, surface normals and height predictions.}
    \end{center}
\end{figure}

This work focuses on the height prediction task that is to predict and reconstruct the corresponding height map, or in other words, predict the height value for every pixel in the input aerial image. Predicting such height maps can be very useful in the subsequent task of 3D reconstruction. By obtaining the accurate height of each building or structure appearing in the input images, 3D models can be generated as an accurate representation of the surrounding world. These 3D models are crucial for GPS-denied navigation, or other fields such as urban planning or telecommunications. Theses reconstructions are traditionally done using Structure from Motion (SfM) \cite{moulon2012adaptive,moulon2013global} technique with stereo camera rigs, which can be very sensible to noise and changes in lighting condition.  \par

For the task of height prediction from aerial images, we propose a multi-task learning framework where additional branches are introduced to improve height prediction accuracy. Previous works have showed that multi-task learning helps improving the accuracy of height prediction networks by including semantic labels \cite{carvalho2019multitask}. We propose to add a third branch to the multi-task network which will be devoted to predicting the surface normals, as shown on Fig. 1. In this configuration, the main height prediction branch will have access to both semantic and geometric guidance, improving the results of the height prediction network. 

However, since the input is only an aerial image, our predictions sometimes can be noisy due to artefacts such as shadows or unexpected changes in color. Therefore, we introduce a refinement network which is a denoising autoencoder taking the outputs from the prediction network, removing the noise present in the prediction and producing a higher quality and more accurate height map. By combining these two steps, we are able to produce results that surpass the current state-of-the-art on multiple datasets. We are also able to produce reasonable semantic labels and surface normal predictions without additional optimizations. \par

In summary, our contributions in this work are the following:

\begin{itemize}

    \item We propose a triple-branch multi-task learning network, including semantic label, surface normal and height prediction.
    \item We introduce a denoising autoencoder as a refinement step for the final height prediction results.
    \item We achieve state-of-the-art performance on two publicly available datasets, and an extensive ablation study shows the importance of each step in the 3D reconstruction pipeline.
    \item We show through two applications how our height prediction pipeline can be used to reconstruct dense 3D point clouds with semantic labels.

\end{itemize}

\section{Related work}

\textbf{Multi-task learning: } This learning framework aims at optimizing a single neural network that can predict multiple related outputs, each represented by a task-specific loss function \cite{caruana1997multitask}. Lately, this approach has become increasingly popular, especially in the area of autonomous driving cars, where multiple outputs (such as object detection, semantic segmentation, motion classification) are derived simultaneously from the input of camera images \cite{teichmann2018multinet,Zhou_2020_CVPR}. \par

\textbf{Height prediction from aerial images: } This task has received a considerable amount of attention by the deep learning and remote sensing communities, especially after the use of UAVs to collect aerial images has become widely accessible. The goal here is to generate a height value for each pixel in an input aerial image. In works such as \cite{amirkolaee2019height},\cite{ghamisi2018img2dsm},\cite{ liu2020im2elevation}, deep learning methods such as residual networks, skip connections and generative adversarial networks are leveraged in order to predict the expected height maps. \par 
Other works such as \cite{carvalho2019multitask, srivastava2017joint} proposed to reformulate the task as a multi-learning problem, by introducing neural networks capable of predicting both the height maps and the semantic labels simultaneously. These works showed that both outputs can benefit from each other, during the simultaneous optimization process of the multi-task network. We choose to extend that formulation by including a third branch in our network tasked for predicting surface normals, which was inspired by previous works \cite{dharmasiri2017joint,eigen2015predicting} in the depth prediction task for autonomous driving cars. Surface normals are also known to be extremely useful during 3D reconstruction tasks and are required for surface and mesh reconstruction algorithms such as the Poisson surface reconstruction algorithm \cite{kazhdan2006poisson} or the Ball pivoting algorithm \cite{bernardini1999ball}. \par

\textbf{Denoising Autoencoders: } Removing noise from images is a traditional task in computer vision. Over the years, many techniques were presented in the literature which can be broadly divided into two categories \cite{fan2019brief} : spatial filtering methods and variational denoising methods. The spatial filtering methods can either be linear, such as mean filtering \cite{gonzalez2004digital} or Wiener filtering \cite{jain1989fundamentals,benesty2010study}, or nonlinear such as median filtering \cite{pitas2013nonlinear} or bilateral filtering \cite{paris2009bilateral}. These filtering methods work reasonably well but are limited. If the noise level becomes too high, these methods tend to lead to over-smoothing of the edges that are present in the image. On the other hand, in variational denoising methods, an energy function is defined and minimized to remove the noise, based on image priors or the noise-free images. Some popular variational denoising methods include total variation regularization \cite{rudin1992nonlinear}, non-local regularization \cite{gilboa2009nonlocal} and low-rank minimization \cite{markovsky2012low}.  \par
Lately, a new trend based on deep learning autoencoders has shown great potential on image denoising. Autoencoder is a class of popular neural networks that has shown to be very powerful across multiple tasks such as segmentation of medical imagery \cite{ronneberger2015u}, decoding the semantic meaning of words \cite{liou2014autoencoder} or solving facial recognition challenges \cite{hinton2011transforming}. For our task, the most useful type of autoencoders available in the literature is the denoising autoencoder. As shown in \cite{vincent2010stacked}, autoencoders can be trained to remove noise from an arbitrary input signal such as an image. We propose to use denoising autoencoder to refine the height predictions from the multi-task learning network. \par

\section{Method}

\subsection{Problem setup } 

Our main objective is to predict an accurate height map using only a monocular aerial image as input. We attempt to do so by constructing a two-stage pipeline, where two different networks are cascaded in serial. The first stage of our pipeline is a multi-task learning network, where the main branch is tasked with predicting preliminary height images, aided by semantic and surface normal information that was extracted by two additional branches of the neural network. The second stage can be seen as a denoising autoencoder: All the predictions from the multi-task network are concatenated and fed into the autoencoder, in order to deal with noisy areas remaining in the height results from the first stage. This effectively produces sharper images that are closer to the ground truth. An overview of the full pipeline can be seen in Fig. 3. 

Fundamentally, the height prediction task is a non-linear regression problem that can be formulated as:

\begin{equation}
\label{eqn:goal}
    \min_{\psi\in\Psi} \sum_i\ell(\mathbf{y}_i, \psi(\mathbf{x}_i))
\end{equation}

where $\psi:\mathcal{X}\rightarrow\mathcal{Y}$ denotes the height prediction mapping function from the feasible space $\Psi$, $\ell:\mathcal{Y}\times\mathcal{Y}\rightarrow\mathbb{R}$ denotes a loss function such as the least-square, $\mathbf{x}_i$ is the input aerial image and $y_i$ is the output height map.

Predicting height only using a single branch neural network is possible. However, previous works such as \cite{carvalho2019multitask, srivastava2017joint} showed that including additional branches to predict other related information such as segmentation labels can be beneficial for both tasks. In our case, in addition to predicting the height maps, we also predict semantic labels and surface normals, which provide semantic and geometric guidance by augmenting the main height prediction branch with information from the semantics and surface normal branches. More details can be found in the height prediction section below. 
Hence, our $\psi$ function can now be defined as:

\begin{equation}
  \psi(\mathbf{x}_i)=\{\mathbf{P}_h,\mathbf{P}_s,\mathbf{P}_n\}
\end{equation}
where $\mathbf{P}_h$, $\mathbf{P}_s$ and $\mathbf{P}_n$ are the height, semantic and surface normal predictions respectively, that are trying to approximate $\mathbf{y}_i=\{\mathbf{P}_h^\ast,\mathbf{P}_s^\ast,\mathbf{P}_n^\ast\}$ where $\mathbf{P}_h^\ast,\mathbf{P}_s^\ast$ and  $\mathbf{P}_n^\ast$ are the height, semantic and surface normal ground truth respectively. Finding a good approximation of the $\psi$ function can be seen as the first stage in our proposed method.

Regression problems such as the one we are facing are difficult to solve due to the high number of values expected to be predicted. This makes our height prediction $\mathbf{P}_h$ noisy by definition, so the use of denoising autoencoders is appropriate in this situation. 

First, we can write: $\mathbf{P}_h=\mathbf{P'}_h+e$ where $\mathbf{P'}_h$ is the clean height value, and $e$ the noise inherent to our approximation of the function $\psi$. By introducing a denoising autoencoder, we can approximate the noise function $\gamma$ such as $\mathbf{P}_h=\mathbf{P'}_h + \gamma(\mathbf{{z}_i})$, where $\mathbf{{z}_i}$ is the concatenation of the outputs of $\psi$ with the input aerial image $x_i$. This makes it possible to re-write equations $(2)$ as $\psi(\mathbf{x}_i)=\{\mathbf{P'}_h + \gamma(\mathbf{{z}_i}),\mathbf{P}_s,\mathbf{P}_n\}$. We can also now define the objective of the second stage of our method such as:

\begin{equation}
  \min_{\gamma\in\Gamma} \sum_i\ell(\mathbf{P}_h^\ast, \mathbf{P}_h - \gamma(\mathbf{{z}_i}))
\end{equation}

In this paper, our goal is to approximate both function $\psi$ and $\gamma$ by using two cascaded deep neural networks.

\subsection{Height prediction network}

We solve the height prediction problem via multi-task learning where, in addition to the main height prediction, semantic and surface normals predictions are conducted too. We found that by re-routing the information in the semantic and surface normal branches to the main height branch, our neural network can learn to predict more accurate height values, especially around the edges. \par 

\begin{figure}[h]
    \begin{center}
        \includegraphics[width=0.63\linewidth]{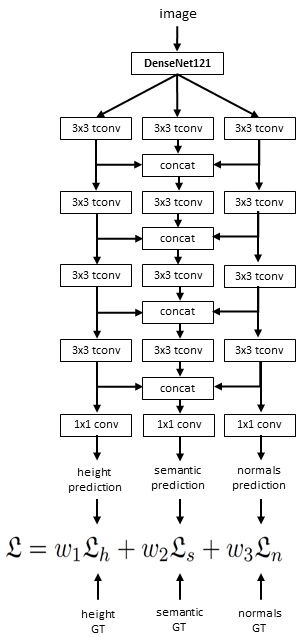}
        \caption{Architecture of our multi-task learning network for height, semantic and surface normals predictions. Note that each tconv block is followed by the ReLu function and drop out layers are inserted after each tconv layers in the main height prediction branch.}
    \end{center}
\end{figure}

Fig. 2 shows our multi-task learning network architecture. We propose a convolutional neural network where we combine a pretrained encoder (tasked with extracting relevant features from the input aerial images), with three inter-connected decoder branches, one for each type of predictions respectively. We chose to use a DenseNet121 network, pretrained on ImageNet, as our main encoder. We show later in the experimentation section that DenseNet121 yields the best accuracy when compared to other popular architectures. Our decoders on the other hand is inspired by \cite{laina2016deeper} and are characterized by being able to reconstruct the expected predictions efficiently. We list in Table 1 the different layers that we used. \par

\begin{figure*}[t]
    \begin{center}
        \includegraphics[width=0.98\linewidth]{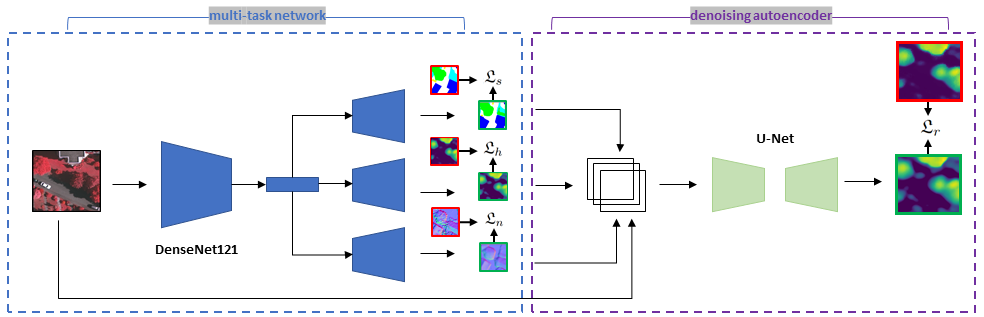}
        \caption{Our two stage height prediction and refinement pipeline. We use DenseNet121 to extract a global feature vector from the input aerial images, which is used to predict the normals map, semantic labels and a first guess at the height map (first stage, in blue). These results are concatenated with the input aerial image and fed into a denoising autoencoder to generate the refined final height map (second stage, in purple). Red boxes represent the ground truth, while green ones represent the networks predictions.}
        \label{fig:net1}
    \end{center}
\end{figure*}

This network is optimized by using a multi-objective loss function defined as:

\begin{equation}
    \mathfrak{L} = w_1\mathfrak{L}_h + w_2\mathfrak{L}_s + w_3\mathfrak{L}_n
\end{equation}

where $\mathfrak{L}_h = \frac{1}{n}\sum_{i=1}^{n}(P_h - P_h^\ast)^{2}$, $\mathfrak{L}_s = -\frac{1}{n}\sum_{i=1}^{n}P_s^\ast log(P_s)$, $\mathfrak{L}_n = \frac{1}{n}\sum_{i=1}^{n}(P_n - P_n^\ast)^{2}$ and $w_1$, $w_2$ and $w_3$ are weights set up according to the training dataset and the scale of each loss function: We found that by using weights that keep all the loss functions at the same scale, the CNN would converge faster and achieve higher final accuracy levels.

\subsection{Height refinement network }
As mentioned previously, the height prediction map $\mathbf{P}_h$ produced by the multi-task learning network still contains some noisy areas that must be refined in order to generate the final height prediction $\mathbf{P'}_h$. We introduce an autoencoder to estimate the noise and produce more accurate height map predictions. \par 
We choose the popular U-Net architecture \cite{ronneberger2015u} as network structure. The input of the network is the concatenation of the multi-task network outputs $\mathbf{P}_h, \mathbf{P}_s$ and $\mathbf{P}_n$  with the aerial image $\mathbf{x}_i$, as shown in Fig. 3. Details of the different layers forming the denoising network are listed in Table 2. The loss function used to optimize this network is the mean square error between the refined height map and the ground truth :  $\mathfrak{L}_r = \frac{1}{n}\sum_{i=1}^{n}(P'_h - P_h^\ast)^{2} =  \frac{1}{n}\sum_{i=1}^{n}(P_h - \gamma - P_h^\ast)^{2}$, with $\gamma$ being the noise function defined in Eq. 3.

\begin{table}[h]
  \begin{center}
    \caption{Height prediction network details. }
\begin{tabular}{|c|c|c|}
    \hline
      &   Layer  &   Output Size\\
    \hline
    {Encoder} & DenseNet121 & (10,10,1024) \\
    \hline
    {Decoder} & $DeConv_1$ & (20,20,1024) \\
     & $Concat$ & (20,20,3072) \\ 
     & $Conv_{11}$ & (20,20,1024) \\
     & $Conv_{12}$ & (20,20,1024) \\
    \cline{2-3}
      & $DeConv_2$ & (40,40,512) \\
     & $Concat$ & (40,40,1536) \\
        & $Conv_{21}$ & (40,40,512) \\
        & $Conv_{22}$ & (40,40,512) \\
    \cline{2-3}
      & $DeConv_3$ & (80,80,256) \\
     & $Concat$ & (80,80,768) \\
        & $Conv_{31}$ & (80,80,256) \\
        & $Conv_{32}$ & (80,80,256) \\
    \cline{2-3}
      & $DeConv_4$ & (160,160,64) \\
     & $Concat$ & (160,160,192) \\ 
        & $Conv_{41}$ & (160,160,64) \\
        & $Conv_{42}$ & (160,160,64) \\
    \cline{2-3}
      & $DeConv_5$ & (320,320,32) \\
     & $Concat$ & (320,320,96) \\ 
        & $Conv_{51}$ & (320,320,32) \\
        & $Conv_{52}$ & (320,320,32) \\
    \cline{2-3}
        & $Conv_{out}$ & (320,320,1) \\
    \hline
    \end{tabular}
  \end{center}
\end{table}

\begin{table}[h]
  \begin{center}
    \caption{Height refinement network details. }
\begin{tabular}{|c|c|c|}
    \hline
      &   Layer   &   Output Size\\
    \hline
    {Encoder} & $Conv_1$ & (320,320,64) \\
     & $MaxPooling$ & (160,160,64) \\
    \cline{2-3}
      & $Conv_2$ & (160,160,128) \\
     & $MaxPooling$ & (80,80,128) \\
    \cline{2-3}
      & $Conv_3$ & (80,80,256) \\
     & $MaxPooling$ & (40,40,256) \\
    \cline{2-3}
        & $Conv_4$ & (40,40,512) \\
     & $MaxPooling$ & (20,20,512) \\
    \cline{2-3}
        & $Conv_5$ & (20,20,1024) \\
    \hline
    {Decoder} & Upsampling & (40,40,512) \\
     
     & $Concat$ & (40,40,1024) \\
      & $Conv_6$ & (40,40,512) \\
    \cline{2-3}
     & $Upsampling$ & (80,80,256) \\
     & $Concat$ & (80,80,512) \\
      & $Conv_7$ & (80,80,256) \\
    \cline{2-3}
     & $Upsampling$ & (160,160,128) \\
     & $Concat$ & (160,160,256) \\
        & $Conv_8$ & (160,160,128) \\
    \cline{2-3}
     & $Upsampling$ & (320,320,64) \\
     & $Concat$ & (320,320,128) \\
        & $Conv_8$ & (320,320,64) \\
    \cline{2-3}
        & $Conv_{out}$ & (320,320,1) \\
    \hline
    \end{tabular}
  \end{center}
\end{table}

\section{Experiments}

\subsection{Datasets }
\textbf{2018 DFC \cite{xu2019advanced} } dataset was released during the 2018 Data Fusion Contest organized by the Image Analysis and Data Fusion Technical Committee of the IEEE Geoscience and Remote Sensing Society. It was collected over the city of Houston, which contains multiple optical resources geared toward urban machine learning tasks such multispectral LiDAR, hyperspectral imaging, Very High-Resolution (VHR) imagery and semantic labels. Using the results of the multispectral LiDAR, it is possible to obtain Digital Structural Models (DSM) and Digital Elevation Models (DEM), which, if subtracted from one another, produces height maps that we can use as ground truth. Four tiles of data are used for training while ten tiles are used for testing.

\textbf{ISPRS Vaihingen \cite{cramer2010dgpf} } dataset was released during the semantic labeling contest of ISPRS WG III/4. It was collected over the city of Vaihingen, Germany and consists of very high resolution true ortho photo (TOP) tiles, corresponding Digital Surface Models (DSM) and semantic labels. As it is usually done when dealing with this dataset, we use the normalized DSM (nDSM) produced by \cite{gerke2014use} as ground truth for our height prediction. Sixteen tiles were used for training while seventeen tiles are used for testing.

\textbf{Surface normal maps:} The surface normal maps for both dataset are generated using the given height maps, following practices usually used for surface normal estimation from dense depth maps based on the Sobel operator \cite{articleSobel}. The details are listed in Alg 1.

\begin{algorithm}
    \SetKwInOut{Input}{Input}
    \SetKwInOut{Output}{Output}

    % \underline{function genNorms} $(P_{h})$\;
    \Input{Height map $P_{h}$}
    \Output{Surface normals map $P_{n}$}
    $zx \gets Sobel(P_{h}, 0)$ \\
    $zy \gets Sobel(P_{h}, 1)$ \\
    $N \gets stack(-zx,-zy,1)$ \\
    $P_{n} \gets \frac{N / \|N\|}{2} + 1$\\
    \textbf{return} $P_{n}$
    
    \caption{Surface normals generation}
\end{algorithm}

\begin{figure}[h]
        \begin{center}
        \includegraphics[width=0.98\linewidth]{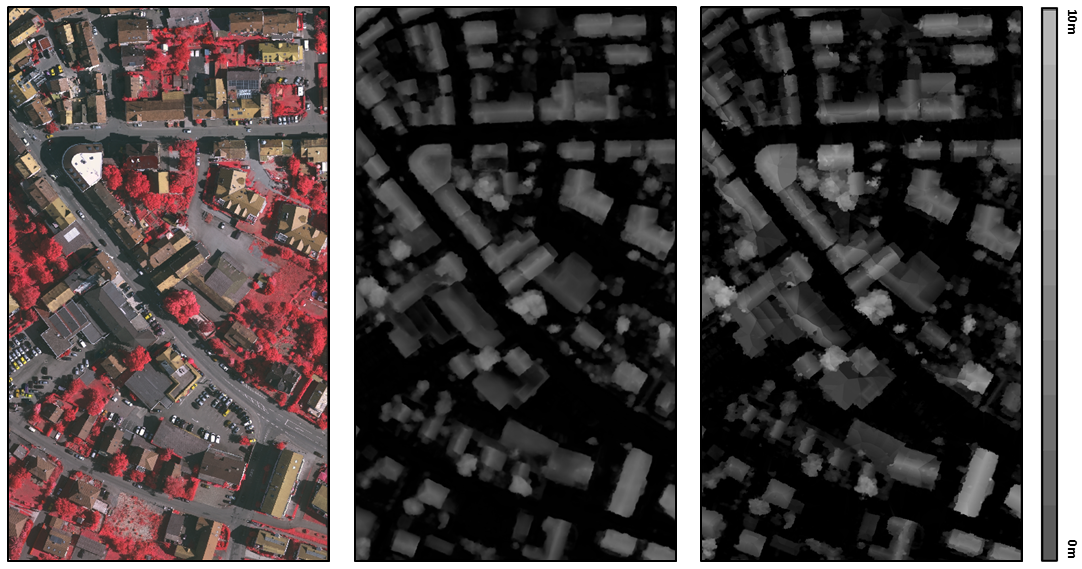}
        \caption{Qualitative comparison of a reconstructed tile from the testing dataset. From left to right: The input RGB tile, the height prediction and the height ground truth.}
    \end{center}
\end{figure}

\subsection{Network Training and Results }

\textbf{Training :} Our training process is not end-to-end. Instead, we follow a two stages approach: we first remove the denoising autoencoder and only focus on training the multi-task network. To do so, random 320x320 crops are sampled from the aerial tiles and corresponding semantic, surface normals and height ground truth are used for training. Once the multi-task network converges, we freeze its weights and then plug into the denoising autoencoder to obtain the final height predictions. We train this second network following the same random sampling process used to train the first one. We use Tensorflow \cite{abadi2016tensorflow}, a learning rate of 0.0002, a batch size of 64, the Adam optimizer\cite{kingma2014adam} and a single RTX2080Ti to train both stages. During training, we saw that altering the network's hyper parameters can sometimes have a slight effect of the convergence speed, but no significant effect on the final accuracy level. \par
Note that in the case of the DFC2018 dataset, the input VHR aerial tiles are ten times bigger than their corresponding DSM, DEM and semantic labels. To deal with that, we first down sample the aerial tiles ten times before starting to collect training crops. \par
\textbf{Results :} The aerial tiles were reconstructed using a sliding window of the same size as of the training samples and with a constant step size. We use Gaussian smoothing to deal with overlapping areas. This makes it possible to deal with cases where different crops of the same area produce different height values, while also protecting the final result from the "checkerboard effect". We report the results of our height prediction and refinement pipeline on both datasets in Table 3, where we use the mean square error (MSE), the mean absolute error (MAE) and root-mean-square error (RMSE) as metrics, all in meters. We also show a qualitative comparison in Fig. 4. When comparing with previous proposed methods in the literature, we can see that by using our multi-task network combined with the refinement step, we are able to surpass the state-of-the-art performance across all metrics on both datasets, with improvement up to 25\%. \par
We credit this increase in accuracy to multiple factors. Firstly, the choice of our encoder (in this case DenseNet121), which is capable of extracting features that are relevant to this task. The second is the context information brought by our 2 additional branches in the multi-task prediction network. Knowing if a pixel falls on a building rather than the road, in addition to the orientation of its associated surface normal vector, helps the network predict height values better. Finally, the denoising autoencoder helps us deal with certain artefacts that tend to confuse the prediction network. We provide numerical analysis of these observations in the ablation study. \par
It is also interesting to note that we are able to achieve similar scores to methods which were trained on the high-definition aerial tiles directly without any down sampling as shown in Table 4. For reconstruction of the same sized area, such networks would take much longer processing time and significantly more computing resources than our proposed method. \par
Missing values in Table 3 were not reported by the cited publications. We also exclude the results reported by \cite{liu2020im2elevation} because it did not follow the same training/testing split of the data.

\begin{table}[h]
  \begin{center}
    \caption{Comparison with other height prediction methods on the ISPRS Vaihingen and the 2018 DFC datasets in meters. }
    \tabcolsep=0.11cm
\begin{tabular}{| *{10}{c|} }
    \hline
 & \multicolumn{3}{c|}{ISPRS Vaihingen} & \multicolumn{3}{c|}{2018 DFC}\\
                            
    \hline
     Method  &   MSE  &   MAE  &   RMSE  &   MSE  &   MAE  &   RMSE\\
    \hline
    Ours   &  \textbf{0.0042}  &   \textbf{0.036}  &   \textbf{0.062}  &  \textbf{6.92}  &   \textbf{1.37}  &  \textbf{2.57}  \\
    \hline
    Carvalho \cite{carvalho2019multitask}   &  0.0060  &   0.045  &   0.074  &   9.34  &   1.53  &  2.97\\
    \hline
    Srivastava \cite{srivastava2017joint}   &  -  &   0.063  &   0.098  &   -  &   -  &  -\\
    \hline
    IMG2DSM \cite{ghamisi2018img2dsm}   &  -  &   -  &   0.090 &   -  &   -  &  - \\
    \hline
    \end{tabular}
  \end{center}
  
  \begin{center}
\caption{Comparison with method trained on VHR aerial images.}
\resizebox{\columnwidth}{!}{%
\begin{tabular}{|c|c|c|c|c|c|}  
    \hline
    Method & MSE & MAE & RMSE & Time (s) & Input Resolution\\
    \hline
    Ours & \textbf{6.92}  &   1.37  &   \textbf{2.57} &  72 & 1192x1202\\
    \hline
    Carvalho VHR \cite{carvalho2019multitask} & 7.27  &  \textbf{1.26}  &  2.59 & 774 &  11920x12020\\
    \hline
\end{tabular}
}
\end{center}

\end{table}

\subsection{Semantic label and surface normal predictions }

Although this work does not focus on the semantic label and surface normal predictions and only uses them to improve the height predictions, we share the results of those two branches and compare them with available methods in the literature in Table 5. Our results in Table 5 show that our multi-task network is able to produce semantic label results that are comparable with the state of the art on the Vaihingen dataset and acceptable ones on the DFC2018 (which has 20 classes compared to the 6 of the Vaihingen dataset). We use the following metrics for the semantic segmentation: The overall accuracy (OA), defined as the sum of accuracies for each class predicted, divided by the number of class, the average accuracy (AA), defined as the number of correctly predicted pixels, divided by the total of pixels to predict and Cohen's coefficient (Kappa), which is defined as $\mathfrak{Kappa} = \frac{p_0 - p_e}{1-p_e}$, such as $p_e$ is the probability of the network classifying a pixel correctly and $p_0$ is the probability of the pixel being correctly classified by chance. The network is also able to produce meaningful surface normal maps as seen on Fig. 1. Missing values in Table 5 were not reported by the cited publications.

\begin{table}[h]
  \begin{center}
    \caption{Semantic labels and surface normals results on the ISPRS Vaihingen and the 2018 DFC datasets. }
\resizebox{0.95\columnwidth}{!}{%
\begin{tabular}{| *{7}{c|} }
    \hline
 & \multicolumn{3}{c|}{ISPRS Vaihingen} & \multicolumn{3}{c|}{2018 DFC}  \\
    \hline
 & \multicolumn{6}{c|}{Semantic Labels}  \\
                            
    \hline
     Method  &   OA  &   AA  &   Kappa  &  OA  &   AA  &   Kappa  \\
    \hline
    Ours   &  85.6  &   74.8  &   \textbf{80.1}  &   51.89  &   47.01  &  49  \\
    \hline
    Carvalho \cite{carvalho2019multitask}   &  \textbf{87.7}  &   \textbf{85.4}  &   75.9  &   \textbf{64.70}  &   \textbf{58.85}  &  \textbf{63}  \\
    \hline
    Srivastava \cite{srivastava2017joint}   &  78.8  &   73.4  &   71.9  &   -  &   -  &  -  \\
    \hline
    Cerra \cite{cerra2018combining}   &  -  &   -  &   -  &   58.60  &   55.60  &  56  \\
    \hline
    Fusion-FCN \cite{xu2018multi}   &  -  &   -  &   -  &   63.28  &   -  &  61  \\
    \hline
 & \multicolumn{6}{c|}{Surface Normals}  \\
                            
    \hline
     Method  &   MSE  &   MAE  &   RMSE  &   MSE  &   MAE  &   RMSE  \\
    \hline
    Ours   &  0.0115  &   0.0642  &   0.1066  &   0.0620  &   0.2119  &  0.2572  \\
    \hline

    \end{tabular}
    }
  \end{center}
%   \vspace{-4mm}
\end{table}

\subsection{Ablation study }

\textbf{Height refinement:} To demonstrate the usefulness of the aforementioned  refinement network, we test our method with and without the denoising autoencoder, on both datasets. In Table 6, we compare the results obtained after both experiments and show that the refinement step always produces more accurate height maps, resulting in an increase of up to 16\% in accuracy. By combining the information present in the semantic and surface normal inputs with the initial guess of the height produced by the previous network, the refinement network is able to concentrate on noisy areas where the height values are abnormal and fix them automatically. In addition, we compare our deep learning based denoiser with other popular non-learning denoising algorithms such as Bilateral Filtering (BF) \cite{paris2009bilateral} and Non-local Means (NIM) regularization \cite{gilboa2009nonlocal}.

\begin{table}[h]
  \begin{center}
    \caption{Comparison of our height prediction methods with and without refinement, on the ISPRS Vaihingen and the 2018 DFC datasets in meters. }

    \tabcolsep=0.11cm
\begin{tabular}{| *{10}{c|} }
    \hline
 & \multicolumn{3}{c|}{ISPRS Vaihingen} & \multicolumn{3}{c|}{2018 DFC}\\
                            
    \hline
     Method  &   MSE  &   MAE  &   RMSE  &   MSE  &   MAE  &   RMSE\\
    \hline
    multi-task only   &  0.0045  &   0.043  &   0.065  &   7.36  &   1.50  &  2.64\\
    \hline
    multi-task + BF   &  {0.0046}  &   {0.043}  &   {0.065}  &  {7.27}  &   {1.51}  &  {2.62}  \\
    \hline
    multi-task + NIM   &  {0.0045}  &   {0.043}  &   {0.065}  &  {7.34}  &   {1.48}  &   {2.63}  \\
    \hline
    multi-task + Unet   &  \textbf{0.0042}  &   \textbf{0.036}  &   \textbf{0.062}  &  \textbf{6.92}  &   \textbf{1.37}  &  \textbf{2.57}  \\
    \hline
    
    \end{tabular}
  \end{center}
\end{table}

We also show qualitatively on Fig. 5 that the refinement height maps are much closer to the ground truth and contains less noise than the direct output of the multi-task network.

\begin{figure}[h]
    \begin{center}
        \includegraphics[width=0.90\linewidth]{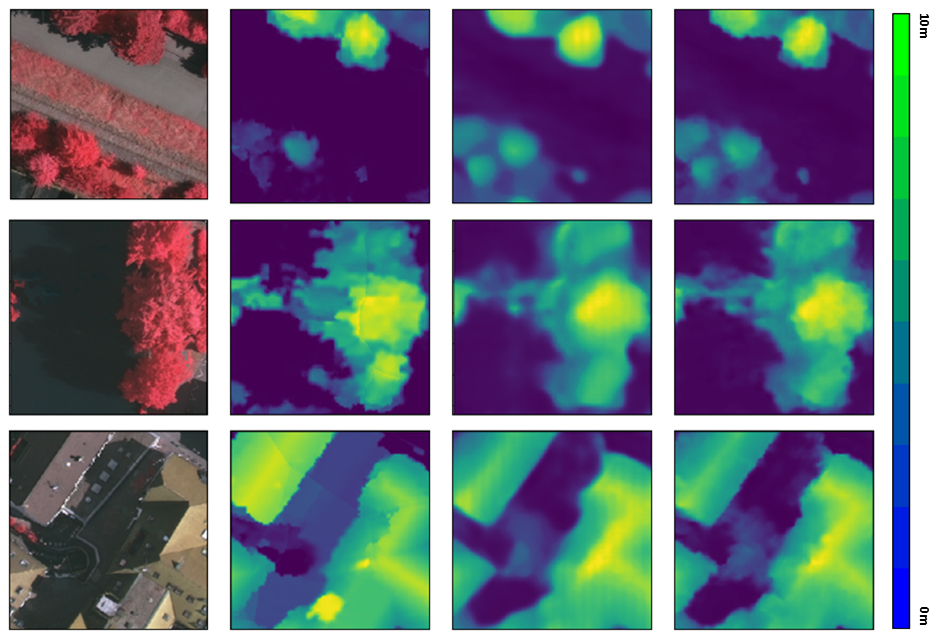}
        \caption{Qualitative comparison. From left to right: The input RGB image, the height prediction of our multi-task network, the refined height map of our denoising autoencoder and the ground truth.}
    \end{center}
\end{figure}

\textbf{Choosing the right encoder :} Our network structure for height prediction is generic, since any off-the-shelf encoder can be used in the first stage to extract features from the input aerial image. 

\begin{table}[h]
  \begin{center}
% \begin{wraptable}{l}{4cm}
\caption{Encoder comparison on the DFC2018 dataset in meters.}
% \begin{center}
\resizebox{0.75\columnwidth}{!}{%
\begin{tabular}{|c|c|c|c|}  
    \hline
    Encoder & MSE & MAE & RMSE\\
    \hline
    ResNet101 \cite{he2016deep} & 18.95 & 3.33 & 4.19 \\
    \hline
    VGG19 \cite{simonyan2014very} & 8.57 & 1.87 & 2.85 \\
    \hline
    DenseNet121 \cite{huang2017densely} & \textbf{7.36} & \textbf{1.50} & \textbf{2.64} \\
    \hline
\end{tabular}
}
% \end{wraptable} 
\end{center}
\end{table}

However, we show in Table 7 that DenseNet121 outperforms other popular encoder structures and produces the most accurate height maps. This is owing to the fact that DenseNet121 is much deeper than the other two networks and contains a higher number of skip connections between layers, making it possible to extract much finer features from the input image. All the networks are trained for the same number of epochs and using the same hyper parameters, such that it ensures the fairness when comparing both the convergence speed and accuracy scores.

\textbf{Geometric and semantic guidance :} In this section, we show the effect of the geometric and semantic guidance in our method in both height prediction and height refinement stages. First, we show in Table 8 that using a multi-task network instead of a single task one improves the overall height prediction results. We also show in Table 9 that by concatenating all the results of the first stage as the input to the denoising autoencoder, we are able to generate more accurate and refined results compared to only using the height image as input. This shows that the semantic and geometric context information brought by two additional branches assist in producing more accurate height values.

\begin{table}[h]
  \begin{center}
    \caption{Comparison of height prediction results of single and multi-task networks in meters. }

\begin{tabular}{| *{7}{c|} }
    \hline
 & \multicolumn{3}{c|}{ISPRS Vaihingen} & \multicolumn{3}{c|}{2018 DFC}  \\
                            
    \hline
     Method  &   MSE  &   MAE  &   RMSE  &   MSE  &   MAE  &   RMSE  \\
    \hline
    single-task   &  0.0048 &   0.046  &   0.067  &   8.17  &   1.64  &  2.78  \\
    \hline
    multi-task   &  \textbf{0.0045}  &   \textbf{0.043} &  \textbf{0.065}  &   \textbf{7.36}  &  \textbf{1.50}  &  \textbf{2.64}  \\
    \hline

    \end{tabular}
  \end{center}
%   \vspace{-4mm}
\end{table}

\begin{table}[h]
  \begin{center}
    \caption{Comparison of height refinement results of single and multi-input denoiser in meters. }

\begin{tabular}{| *{7}{c|} }
    \hline
 & \multicolumn{3}{c|}{ISPRS Vaihingen} & \multicolumn{3}{c|}{2018 DFC}  \\
                            
    \hline
     Method  &   MSE  &   MAE  &   RMSE  &   MSE  &   MAE  &   RMSE  \\
    \hline
    single-input   &  0.0043 &   0.037  &   0.063  &   7.13  &   1.47  &  2.62  \\
    \hline
    multi-input  &  \textbf{0.0042}  &   \textbf{0.036} &  \textbf{0.062}  &   \textbf{6.92}  &  \textbf{1.37}  &  \textbf{2.57}  \\
    \hline

    \end{tabular}
  \end{center}
%   \vspace{-4mm}
\end{table}

\textbf{Finding the right reconstruction step :} The accuracy of our final tile reconstruction depends also on the step size of the sliding window that we choose when collecting the aerial crops. We show in Table 10 the different results corresponding to different step sizes. We found that a step size of 60 pixels results the best across both datasets.

\begin{table}[h]
  \begin{center}
    \caption{Comparison of our reconstruction results (meters) based on the step size (pixels). }

\begin{tabular}{| *{7}{c|} }
    \hline
 & \multicolumn{3}{c|}{ISPRS Vaihingen} & \multicolumn{3}{c|}{2018 DFC}  \\
                            
    \hline
    Step  &   MSE  &   MAE  &   RMSE  &   MSE  &   MAE  &   RMSE  \\
    \hline
    80  &  0.00421  &   0.0363 &  0.0625  &   6.98  &  1.38  &  2.58  \\
    \hline
    60  & \textbf{0.00420}  &   \textbf{0.0362} &  \textbf{0.0623}  &   \textbf{6.92}  &  \textbf{1.37}  &  \textbf{2.57}  \\
    \hline
    40  &  0.00421  &   0.0362 &  0.0623  &   6.93  &   1.37 &  2.58  \\
    \hline

    \end{tabular}
  \end{center}
%   \vspace{-4mm}
\end{table}

\textbf{Visualizing the uncertainty :} In order to investigate the performance of our pipeline more thoroughly, we generate uncertainty maps according to the method proposed in \cite{kendall2015bayesian}. The results are displayed in Fig. 6 and show that most of the prediction errors can be attributed to the areas such as the edges of buildings due to the sudden changes in brightness and color, and trees where shadows introduce a significant amount of color noise.

\begin{figure}[h]
    \begin{center}
        \includegraphics[width=0.85\linewidth]{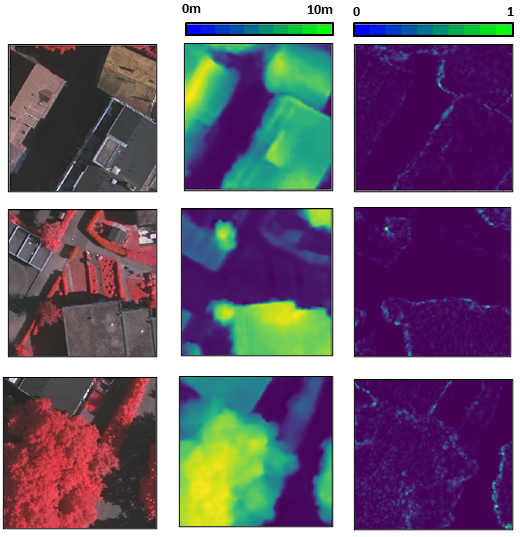}
        \caption{Uncertainty results. From left to right RGB Image, Height Prediction, Uncertainty Map. Prediction errors are mostly concentrated around the edges.}
    \end{center}
\end{figure}

\section{Applications for 3D Reconstruction}

In this section, we propose two applications to show how to take advantage of the results generated by our proposed pipeline. The first is 3D reconstruction of select buildings from a single aerial image. In the second application, we simulate a UAV flight over a certain area and show that we can reconstruct the entire 3D area by combining odometry and aerial images. In comparison to the classic SfM algorithm, our method provides a significant gain in speed, accuracy and density. More importantly, our proposed method requires significantly less number of images since only minimal overlaps are necessary when taking the aerial shots.

\subsection{Single aerial image 3D reconstruction }
Usually, in order to reconstruct the 3D shape of a building, multiple shots from multiple angles with significant overlap are necessary in order to apply the sequential surface from motion algorithm. We show in Fig. 7(b) that owing to our multi-task network, we are able to produce accurate 3D point clouds of the buildings using a single image only. \par
The proposed method is also capable of generating semantic point clouds in Fig. 7(c) and 3D meshes of buildings and their surrounding areas in Fig. 7(d) by leveraging the semantic labels and surface normals generated by the networks. Specifically, semantic point clouds are generated by projecting the semantic labels onto the point clouds, while the meshes are generated by combining the surface normals with the reconstructed point clouds using the ball pivoting algorithm \cite{bernardini1999ball}.

\begin{figure*}

    \begin{center}
    
        \includegraphics[width=0.95\linewidth]{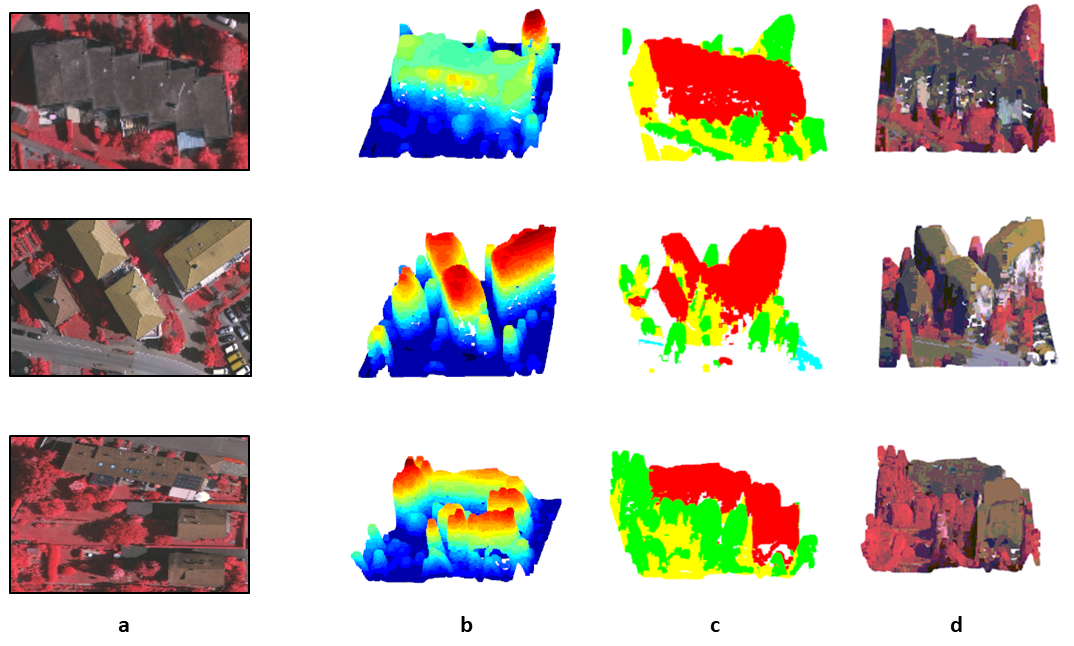}
        \caption{3D reconstructions using a single image. (a) RGB Image, (b) Height Colorized Pointcloud, (c) Semantic Pointcloud, (d) RGB Colorized Mesh.}

    \end{center}
    
\end{figure*}

\begin{figure*}
    \begin{center}
        \includegraphics[width=0.95\linewidth]{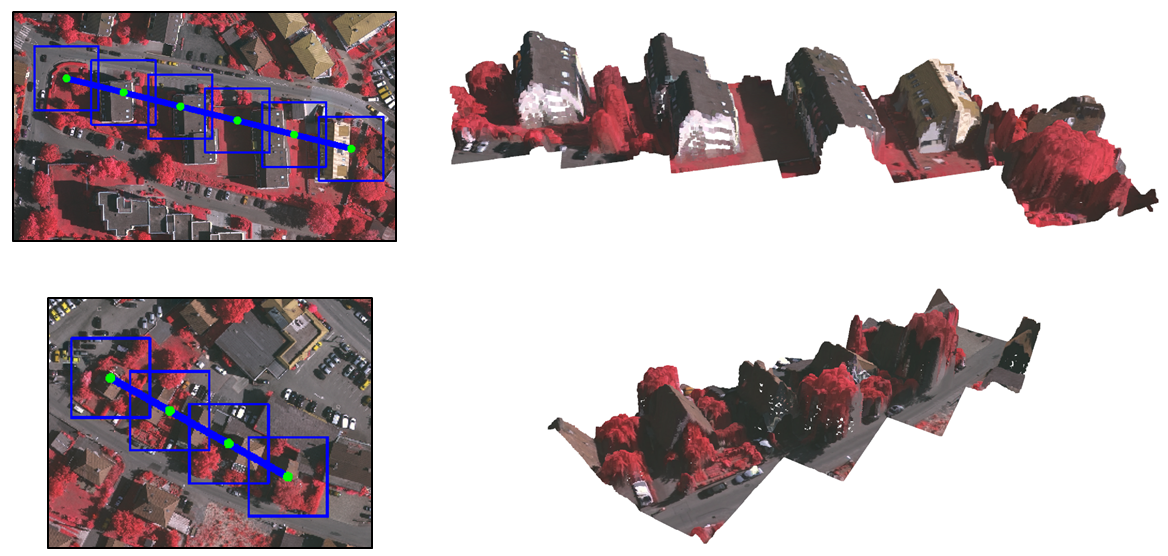}
        \caption{3D reconstructions from simulated UAV flight. From left to right: Positions of the UAV images, Reconstructed 3D scene.}
    \end{center}
\end{figure*}

\subsection{Area reconstruction with simulated UAV flight }
3D reconstruction of urban areas is a very useful application. Similarly to what we mentioned in the first application, reconstructing an entire area would generally require a series of captured images with significant overlaps, by flying the drones in multiple passes over the same area, in order to generate a semi-dense point cloud. \par 
In our case, we show in Fig. 8 that by using a single pass with a small number of captured images and minimal overlap (only to avoid gaps in the final reconstruction) we are able to produce accurate and dense 3D reconstructions. We also note that when we feed the same data to an SfM algorithm, it typically leads to failures since only a small number of features can be matched among the single-pass aerial shots. The data is collected by simulating a constant altitude UAV flight over a certain neighborhood in one of the tiles available in the testing datasets. The odometry is assumed to be known from on-board IMU or GPS sensors.

\section{Conclusion}
In this work, we propose a deep learning based two-stage pipeline that can predict and refine height maps from a single aerial image. We leverage the power of multi-task learning by designing a three-branch neural network for height, semantic label and surface normal predictions. We also introduce a denoising autoencoder to refine the predicted height maps and largely eliminate the noise remaining in the results of the first stage height prediction network.  Experiments on two publicly available datasets show that our method is capable of outperforming state-of-the-art results in height prediction accuracy. In future work, we plan on exploring the computational efficiency of the proposed neural networks for their applications towards real-time processing of aerial images.

% or


\

\ifCLASSOPTIONcaptionsoff
  \newpage
\fi



% http://www.michaelshell.org/tex/bibtex/
\bibliographystyle{IEEEtran}

 \bibliography{egbib}

\begin{IEEEbiography}[{\includegraphics[width=1in,height=1.25in,clip,keepaspectratio]{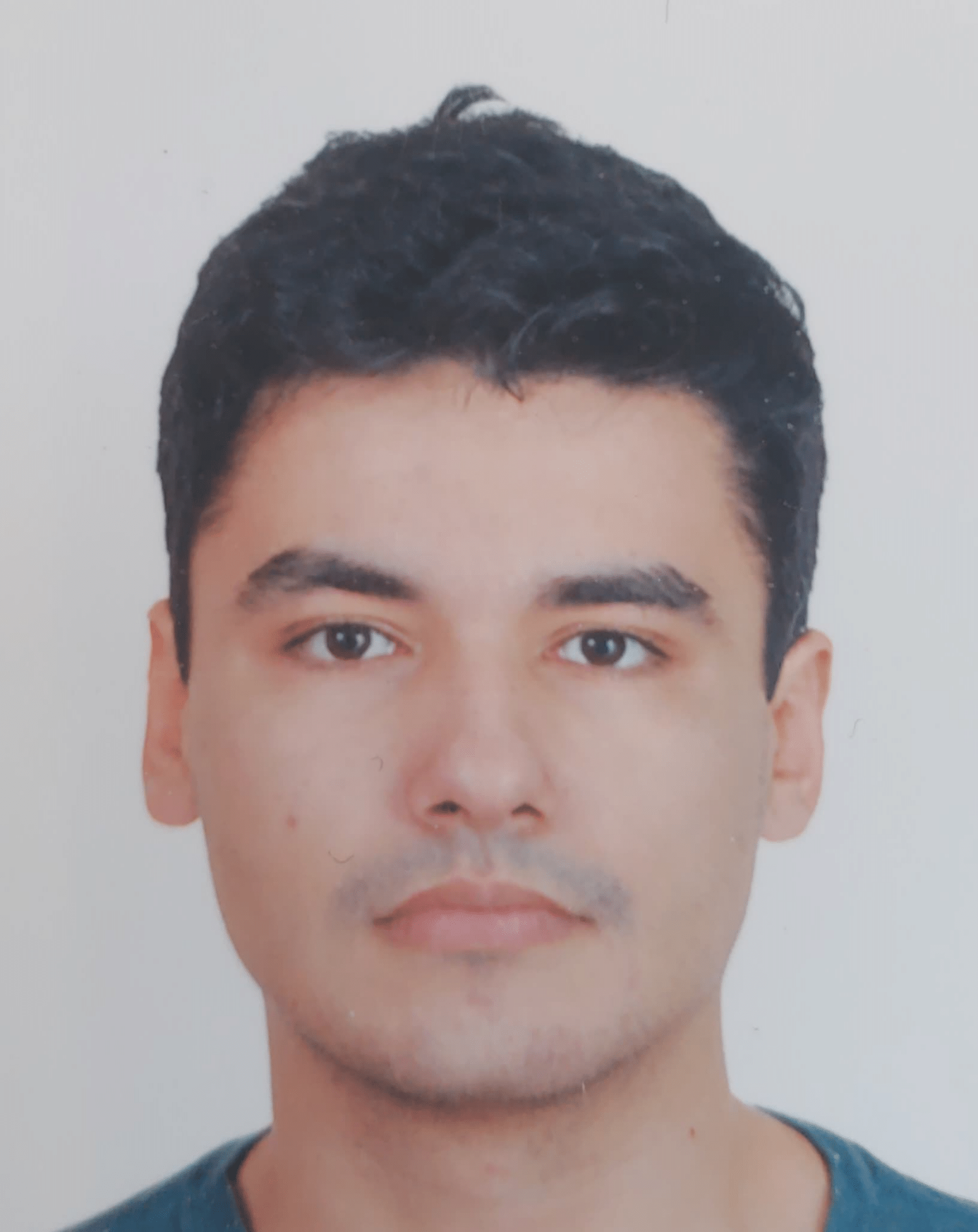}}]{Mahdi Elhousni}

is currently pursuing a PhD in Electrical and Computer Engineering at the Worcester Polytechnic in Worcester, MA, USA. Before joining WPI, he had received a BS in computer science and a MS in embedded systems from the National school For Computer Science in Rabat, Morocco. His main research interest are computer vision, deep learning and SLAM. \par
  
\end{IEEEbiography}

\begin{IEEEbiography}[{\includegraphics[width=1in,height=1.25in,clip,keepaspectratio]{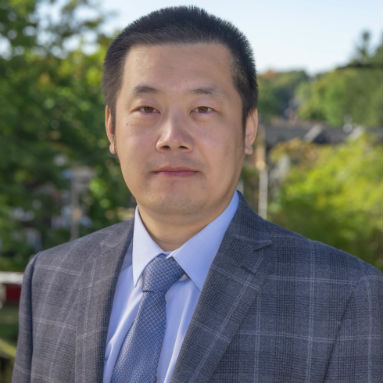}}]{Ziming Zhang} 

is an assistant professor at Worcester Polytechnic Institute. Before joining WPI he was a research scientist at Mitsubishi Electric Research Laboratories (MERL) in 2016-2019. Prior to that, he was a research assistant professor at Boston University. Dr. Zhang received his PhD in 2013 from Oxford Brookes University, UK, under the supervision of Prof. Philip H. S. Torr (now in the University of Oxford). His research areas lie in computer vision and machine learning, especially in object recognition/detection, data-efficient learning (e.g. zero-shot learning) and applications (e.g. person re-identification), deep learning, optimization. His works have appeared in PAMI, CVPR, ICCV, ECCV, NIPS. He serves as a review/PC member for top conferences (e.g. CVPR, ICCV, NIPS, ICML, ICLR, AAAI, AISTATS, IJCAI) and journals (e.g. PAMI, IJCV, JMLR). He won the R\&D100 Award 2018.\par
  
\end{IEEEbiography}

\begin{IEEEbiography}[{\includegraphics[width=1in,height=1.25in,clip,keepaspectratio]{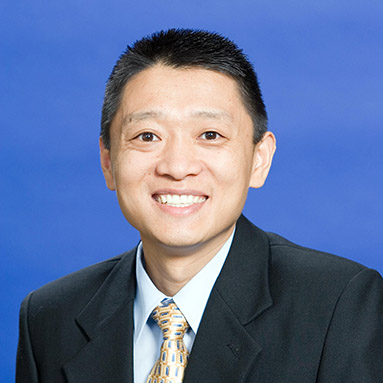}}]{Xinming Huang} 

received his Ph.D. degree in electrical engineering from Virginia Tech, in 2001. He was a Member of Technical Staffs with the Wireless Advanced Technology Laboratory, Bell Labs of Lucent Technologies. Since 2006, he has been a Faculty Member with the Department of Electrical and Computer Engineering, Worcester Polytechnic Institute (WPI), where he is currently a Full Professor. His main research interests include the areas of circuits and systems, with an emphasis on reconfigurable computing, wireless communications, information security, computer vision, and machine learning.\par
  
\end{IEEEbiography}

\EOD
\end{document}